%% file: main.tex
\documentclass{article}

\usepackage[nonatbib, final]{neurips_2019}

\usepackage[utf8]{inputenc} 
\usepackage[T1]{fontenc}    
\usepackage{hyperref}       
\usepackage{url}            
\usepackage{booktabs}       
\usepackage{amsfonts}       
\usepackage{nicefrac}       
\usepackage{microtype}      
\usepackage{enumitem}
\usepackage{graphicx} 
\usepackage{textcomp}
\usepackage{multirow}
\usepackage{wrapfig}

\title{Drug Repurposing for Cancer: An NLP Approach to Identify Low-Cost Therapies}
\vspace{-1cm}
\author{Shivashankar Subramanian\textsuperscript{\rm 1,2} 
\And 
Ioana Baldini\textsuperscript{\rm 1}
\And 
Sushma Ravichandran\textsuperscript{\rm 1} 
\And Dmitriy A. Katz-Rogozhnikov\textsuperscript{\rm 1}
\And
Karthikeyan Natesan Ramamurthy\textsuperscript{\rm 1}
\And 
Prasanna Sattigeri\textsuperscript{\rm 1}
\And 
Kush R. Varshney\textsuperscript{\rm 1}
\And
Annmarie Wang\textsuperscript{\rm 3,4}
\And
Pradeep Mangalath\textsuperscript{\rm 3,5} 
\And
Laura B. Kleiman\textsuperscript{\rm 3}\\
\textsuperscript{\rm 1}IBM Research, Yorktown Heights, NY, USA\\
\textsuperscript{\rm 2}University of Melbourne, VIC, Australia\\
\textsuperscript{\rm 3}Cures Within Reach for Cancer, Cambridge, MA, USA\\
\textsuperscript{\rm 4}Massachusetts Institute of Technology, Cambridge, MA, USA \\
\textsuperscript{\rm 5}Harvard Medical School, Boston, MA, USA
}
\begin{document}

\maketitle
\begin{abstract}
\input{abstract}
\end{abstract}

\input{intro}
\input{problem}
\input{dataset}
\input{results}
\input{conclusion}

\small

\bibliographystyle{abbrv}
\bibliography{redrug}
\end{document}

%% file: abstract.tex
More than 200 generic drugs approved by the U.S. Food and Drug Administration for non-cancer indications have shown promise for treating cancer. Due to their long history of safe patient use, low cost, and widespread availability, repurposing of generic drugs represents a major opportunity to rapidly improve outcomes for cancer patients and reduce healthcare costs worldwide. Evidence on the efficacy of non-cancer generic drugs being tested for cancer exists in scientific publications, but trying to manually identify and extract such evidence is intractable. In this paper, we introduce a system to automate this evidence extraction from PubMed abstracts. Our primary contribution is to define the natural language processing pipeline required to obtain such evidence, comprising the following modules: querying, filtering, cancer type entity extraction, therapeutic association classification, and study type classification. Using the subject matter expertise on our team, we create our own datasets for these specialized domain-specific tasks. We obtain promising performance in each of the modules by utilizing modern language modeling techniques and plan to treat them as baseline approaches for future improvement of individual components.

%% file: intro.tex
\section{Introduction}

Each year around the world, nearly 10 million people die from cancer~\cite{worldcancerstats} and the cost of cancer exceeds USD \$1 trillion~\cite{cancereconomics}. 
Finding new therapeutic uses for inexpensive generic drugs ("drug repurposing") can rapidly create affordable new treatments. Hundreds of non-cancer generic drugs have shown promise for treating cancer, but it is unclear which are the most worthwhile repurposing opportunities to pursue.
Scientific publications such as preclinical laboratory studies and small clinical trials contain evidence on generic drugs being tested for cancer use. The Repurposing Drugs in Oncology (ReDO) project, through manually inspecting research articles indexed by PubMed, found anti-cancer evidence for more than 200 non-cancer generic drugs ~\cite{pantziarka2017repurposing,bouche2017beyond,verbaanderd2017repurposing}. However, manual review to identify and analyze potential evidence is time-consuming and intractable to scale. As PubMed indexes millions of articles and the collection is continuously updated, it is imperative to devise (semi)automated techniques to synthesize the existing evidence. Machine learning (ML)-powered evidence synthesis would provide a comprehensive and real-time view of drug repurposing data and enable actionable insights. To this end, we must achieve task automation, algorithmic accuracy, and technical scalability in three key areas: evidence identification, extraction, and synthesis. 

The work presented in this paper is part of an ambitious initiative to synthesize the plethora of scientific and real-world data on non-cancer generic drugs to identify the most promising therapies to repurpose for cancer. This type of endeavor requires close collaboration between experts in different disciplines, such as cancer research (to provide guidance, annotate datasets, and verify results), machine learning (to devise machine learning tasks, select datasets to be annotated, devise models, and evaluate performance), and software engineering (to incorporate models in end-to-end online applications). Furthermore, implementing repurposed therapies as the standard of care in medical practice requires definitive clinical trials, new incentives and business models to fund them, and engagement by various stakeholders such as patients, doctors, payers, and policymakers. In this paper, we focus on the key aspect of identifying and extracting relevant evidence from PubMed articles.

Methods for synthesizing drug repurposing evidence can be divided into three major categories: network-based methods, natural language processing (NLP) approaches, and semantic techniques~\cite{xue2018review}. Network-based approaches aim to infer relationships between biological entities (drug--disease or drug--target relationships), inspired by the fact that biologic entities (disease, drug, protein, etc.) in the same module of biological networks share similar characteristics~\cite{martinez2015drugnet}. NLP approaches aim to both identify biological entities and mine new knowledge from scientific literature~\cite{li2009building}. Semantic approaches require a semantic network to be built first, which can be used with various approaches to mine relationships between entities~\cite{palma2014drug}. We focus on NLP approaches. Our primary contributions, described in detail in the remainder of this paper, are as follows: formulating the pipeline of NLP tasks required to identify relevant evidence of generic drug repurposing for cancer from PubMed articles, precisely specifying the NLP tasks in terms of input and output (not an easy endeavor), Creating domain-specific datasets that support the task definition, designing and evaluating initial models for each of the domain-specific tasks.



%% file: problem.tex
\section{NLP Pipeline and Dataset for Drug-Cancer Evidence Extraction}
\label{pd}

\begin{figure*}[!tb]
\centering
\includegraphics[scale=0.17]{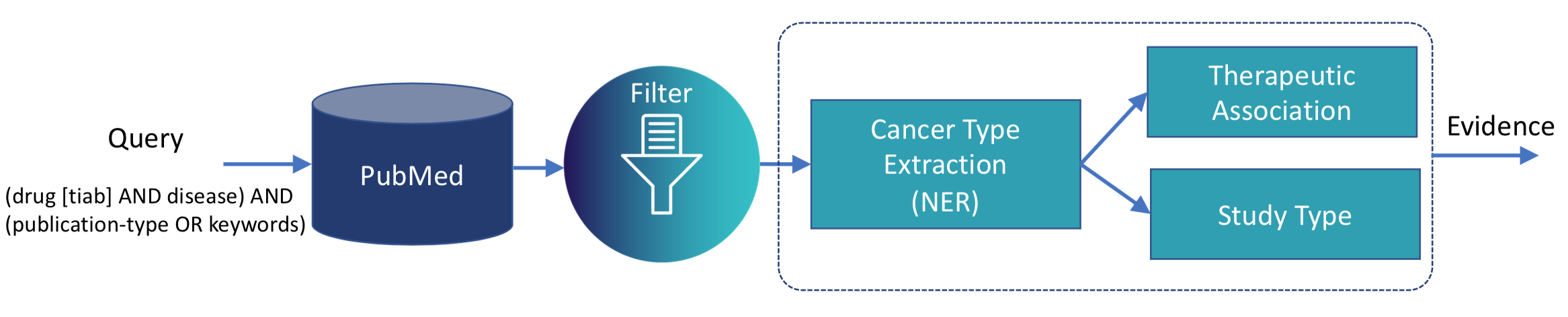}
\caption{Solution overview: Input to the evidence discovery pipeline is a list of non-cancer generic drugs, and the output is published scientific evidence for each drug for treating various cancer types.}
\vspace{-.3cm}
\label{fig:flow}
\end{figure*}


\href{https://www.ncbi.nlm.nih.gov/pubmed/}{PubMed}, provided by the National Center for Biotechnology Information (NCBI), is a comprehensive source of biomedical studies, comprising more than 30 million biomedical abstracts and citations from various sources such as MEDLINE, life science journals, and online books. Given a list of generic drugs, the goal of our work is to automatically select abstracts from the large PubMed collection, that measure cancer-relevant phenotypic outcomes of interventions with generic drugs.\footnote{Phenotype is the observable physical properties of an organism; these include the organism's appearance, development, and behavior.
We focus on phenotypic outcomes (such as proliferation/death of cells grown in culture or tumor progression/overall survival rates for clinical trials) since they are a more direct measure of outcomes that matter to cancer patients and represent stronger therapeutic evidence (as opposed to, for example, the effects of drugs on protein levels).}

We propose an evidence discovery pipeline, shown in Figure \ref{fig:flow}. First we query PubMed using a strategy inspired by the Cochrane highly sensitive search (CHSS) strategy~\cite{dickersin1994systematic} to narrow the collection of articles we analyze. Note that querying PubMed, even with a sophisticated search string, may not yield only \textit{relevant} articles. Hence we have a (shallow) filtering stage to reject the easy irrelevant cases. Using the resulting abstracts, cancer types are identified using a named entity recognition (NER) model. With the abstract and pairs of drug-cancer types, therapeutic association is classified and also the type of study is categorized. We refer to this collection of information (i.e., drug, cancer, therapeutic association, study type) as the \emph{evidence} discussed in the PubMed abstract.

The therapeutic association schema contains the following classes: \textit{1. Irrelevant:} A. Drug has no relation to cancer (cases where either the drug or the cancer is not the focus of the study); B. Abstract does not discuss a \textit{phenotypic outcome} and \textit{2. Relevant:} A. Effective: the drug was shown to be effective for treating the cancer; B. Detrimental: the drug has a detrimental effect on the cancer; C. No effect: the drug has no effect on the cancer; D. Inconclusive: the results of the study are inconclusive.

The study types we consider are defined as follows: preclinical studies (in vitro, in vivo), observational studies (including case reports), and clinical trials. 

\begin{figure}
\centering
  \includegraphics[scale=0.25]{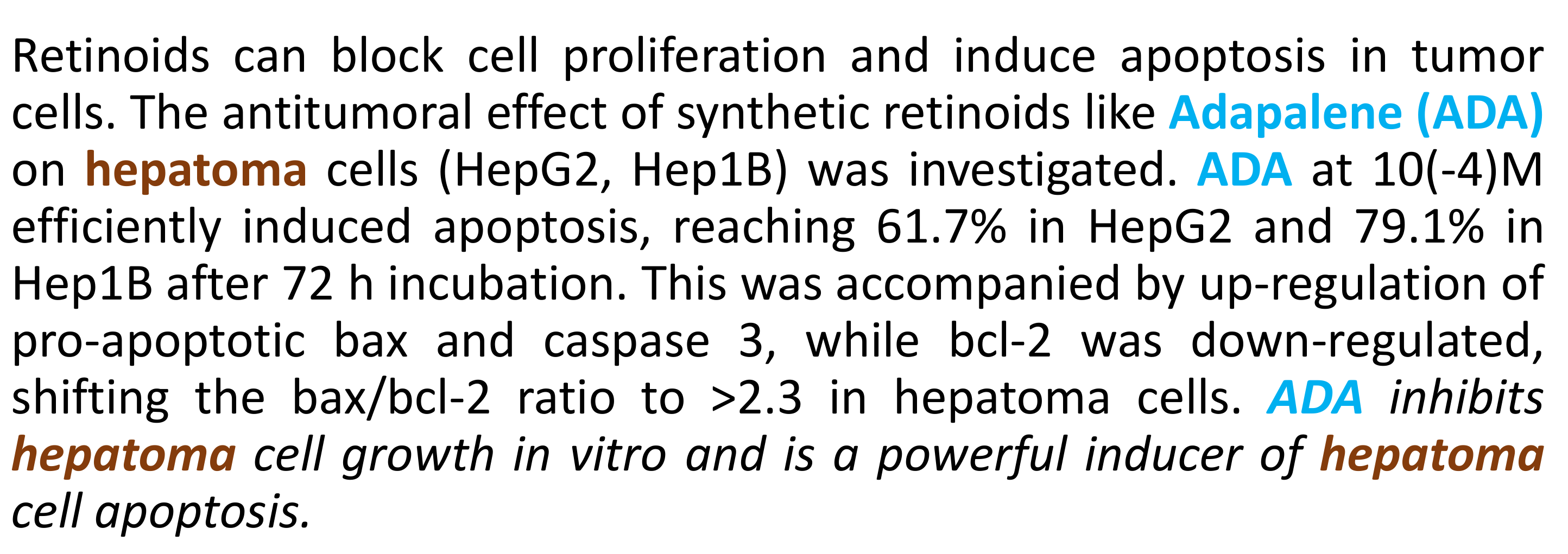}
\vspace{-.3cm}
\caption{Sample \textit{relevant} abstract annotation. PubMed \# 15105045, \emph{Adapalene} is the non-cancer generic drug, used to treat \emph{hepatoma cancer}. It is a preclinical evidence (in vitro study), and has an \textit{effective} association. Evidence for association with phenotypic outcome measured  is italicized.}
\label{fig:ex}
\end{figure}

Identifying such evidence from scientific abstracts is not trivial. The articles that discuss cancer interventions use domain-specific jargon which makes the text hard to comprehend by both humans with non-expert background and machines that are not trained with domain-specific data. Hence a strong collaboration between domain-experts and data scientists is required to define machine learning tasks, collect and annotate the appropriate information and design and evaluate machine learning models that address the designed tasks. Due to space limitations, we cannot elaborate in this paper on all difficulties in manually annotating datasets for all these tasks. We will refer to them during the presentation of our work. An example of an annotated abstract is shown in Figure~\ref{fig:ex}.

%% file: dataset.tex
Our team of machine learning and biomedical scientists worked closely together to fine-tune the querying and filtering strategy and to annotate cancer types, along with the therapeutic associations and study types. In the interest of space, we present the results of the dataset creation in Table~\ref{table:dataset} without the details of the iterative process of producing it. 
\vspace{-.4cm}
\begin{table}
\footnotesize
\begin{minipage}{0.5\textwidth}
\begin{tabular}{l l l }
\toprule
Coarse-level & Association & Count \\
\midrule
\multirow{2}{*}{Irrelevant} & No relation to cancer & 553\\ \\[-0.9em]
& No phenotypic outcome & 155 \\ \\[-0.9em]
\midrule
\multirow{4}{*}{Relevant} & Effective & 555 \\ \\[-0.9em]
&Detrimental & 50 \\ \\[-0.9em]
&Inconclusive & 216 \\ \\[-0.9em]
&No effect & 68 \\ \\[-0.9em]
\bottomrule
\end{tabular}
\end{minipage}
\begin{minipage}{0.5\textwidth}
\begin{tabular}{l l}
\toprule
Study type & Count \\
\midrule
Preclinical & 318 \\
Clinical observational study & 107 \\
Clinical trial & 69\\
Other & 28 \\
\bottomrule
\end{tabular}
\end{minipage}
\caption{Dataset statistics for therapeutic association distribution (left) and study type distribution (right) for the relevant abstracts.}
\vspace{-.7cm}
\label{table:dataset}
\end{table}

%% file: results.tex
\section{Models for Cancer Type, Therapeutic Association, and Study Type}

We briefly discuss the models and their performance for cancer entity extraction, therapeutic association classification, and study type classification, which form the key components of the proposed evidence discovery pipeline.

\subsection{Cancer Type Extraction}
For cancer entity identification, we use two main named entity recognition (NER) methods. We train sequential token-level IOB (inside, outside, beginning) tag prediction model using the BioNLP13CG dataset \cite{pyysalo2015overview}. Tokens that are not of interest are treated as `O'. We use the well-known conditional random field (CRF)~\cite{song2018comparison} and convolutional neural network (CNN) based SpaCy models~\cite{spacy} for entity extraction. We evaluate the performance on the 1085 abstracts using recall\footnote{Ratio between \textit{count of unique cancer entities predicted correctly by the model} and \textit{number of unique cancer entities}.} with exact match and a token-level overlap score \cite{moreau2008robust}, where the predicted entity with highest overlap is used to compute the score. The CRF-based model obtains a recall of 54.2\% and an overlap score of 66.4\%, while the SpaCy-based model presents higher performance of 67.7\% recall and 77.6\% overlap score.




\subsection{Therapeutic Association Classification}

Given a drug-cancer pair and the corresponding abstract text, we build three different models for therapeutic association classification: \textit{1. Logistic Regression} with feature vectors that are a concatenation of term frequency bag-of-words representations of abstracts, drug and cancer type \cite{naacl19};
\textit{2. Deep Averaging Networks (DAN)}: similar to logistic regression, except that the tokens of abstract, drug and cancer type are initialized with word vectors trained using skip-gram objective over a large set of PubMed abstracts~\cite{moendistributional}; the text of a given abstract is passed through deep averaging networks~\cite{iyyer2015deep} where the word vectors are re-trained (with the training data and classification objective), and the representations of abstract, drug and cancer type are concatenated, and passed through a final logistic layer; and \textit{3. SciBERT~\cite{devlin2019bert,beltagy2019scibert}}: The drug and cancer type entities are encapsulated with special characters and concatenated with the input abstract text. The task is framed as a multi-class classification problem. The sequence representation is obtained using SciBERT's encoding of the [CLS] token\footnote{[CLS] is inserted as a special beginning token for every input sequence.} (from the last hidden layer). This encoding captures the entire sequence representation and is used for the multi-class classification with a logistic layer.

We perform a 5-fold cross-validation split at the document level, and evaluate performance for drug-cancer type pairs, given the abstract text. Note that we use gold-standard cancer type annotations for this analysis. We evaluate two different settings to understand the complexity of the task: (1) irrelevant vs.\ relevant binary classification (Table \ref{tab:bin}) and (2) all six classes (Table \ref{table:results}, left side). Performance is measured using F-score. SciBERT performs the best in most cases.

\subsection{Study Type Classification}
We train logistic regression models with different choices of features \cite{marshall2018machine}: bag-of-words (BoW), publication type (PT), MeSH terms and combining all. Results using logistic regression with different choices of features are given in Table \ref{table:results} (right side). Performance is measured using F-score. Using all the features together provides the best performance.

\begin{table}
\centering
\scalebox{0.8}{
\begin{tabular}{l l l l}
\toprule
Class & Log. Reg & DAN & SciBERT \\
\midrule
Irrelevant & \textbf{0.83} & 0.81 & 0.81 \\
Relevant  & 0.80 & 0.79  & \textbf{0.85} \\
\bottomrule
\end{tabular}
}
\caption{Binary therapeutic association classification}
\label{tab:bin}
\end{table}

\begin{table}
\footnotesize
\begin{minipage}{0.5\textwidth}
\scalebox{0.8}{
\begin{tabular}{l l l l}
\toprule
Class & Log. Reg & DAN & SciBERT \\
\midrule
No relation & 0.74 & 0.71 & \textbf{0.80} \\
No phenotypic outcome & 0.30 & 0 & \textbf{0.34}\\
Effective  & 0.72 & 0.67  & \textbf{0.73} \\
Detrimental  & 0.12 & 0  & \textbf{0.33}\\
No effect  & \textbf{0.18} & 0  & \textbf{0.18}\\
Inconclusive  & \textbf{0.30} & 0.03  & 0.25\\
\bottomrule
\end{tabular}
}
\end{minipage}
\begin{minipage}{0.5\textwidth}
\scalebox{0.8}{
\begin{tabular}{l l l l l}
\toprule
Class & PT & BoW & MeSH & All \\
\midrule
Preclinical & 0.86 &\textbf{0.96} & 0.95 & \textbf{0.96}\\
Clinical observational study  & 0.25 & 0.81  & 0.66 & \textbf{0.84}\\
Clinical trial  & 0.78 & 0.78  & 0.60 & \textbf{0.80}\\
Other & 0.38 & 0.34 & 0.37 & \textbf{0.40}\\
\bottomrule
\end{tabular}
}
\end{minipage}
\caption{Fine-grained therapeutic association classification (left) and Study type classification using logistic regression (right).}
\vspace{-.7cm}
\label{table:results}
\end{table}




%% file: conclusion.tex
\section{Conclusion and Future Work}
We proposed an end-to-end evidence discovery pipeline that fetches potential candidate abstracts from PubMed for further evaluation with the goal of identifying non-cancer generic drug activity against different cancer types. We discuss the components in the pipeline, and use NLP approaches along with a number of well-thought-out heuristics to provide solutions for each component.